\documentclass[final,5p,times,twocolumn]{elsarticle}

\usepackage{amssymb}
\usepackage{booktabs}
\usepackage{adjustbox}
\usepackage{siunitx}
\usepackage{lscape}
\usepackage{multirow}
\usepackage{tikz}
\usepackage{colortbl}
\usepackage{array}
\usepackage{svg}
\usepackage{graphicx} 
\usepackage{subfig}
\usepackage{array}
\usepackage{comment}
\usepackage[unicode,breaklinks=true]{hyperref}
\hypersetup{
	colorlinks=true,
   	linkcolor=blue,
   	citecolor=blue,
    filecolor=blue,
   	urlcolor=blue}
\usepackage{colortbl}
\usepackage{hhline}
\usepackage{pdflscape}
\usepackage{caption} 
\usepackage{graphicx,calc}
\usepackage{makecell}

\newcolumntype{L}[1]{>{\raggedright\let\newline\\\arraybackslash\hspace{0pt}}m{#1}}
\newcolumntype{C}[1]{>{\centering\let\newline\\\arraybackslash\hspace{0pt}}m{#1}}
\newcolumntype{R}[1]{>{\raggedleft\let\newline\\\arraybackslash\hspace{0pt}}m{#1}}

\usepackage{soul}
\usepackage{color}
    \definecolor{celadon}{rgb}{0.67, 0.88, 0.69}
    \definecolor{flamingopink}{rgb}{0.99, 0.56, 0.67}
    \definecolor{lovelygreen}{rgb}{0.54, 0.90, 0.60}

\usepackage[figuresright]{rotating}
\usepackage{pbox}
\usepackage{makecell}
\usepackage{appendix}

\usepackage{footnote}
\usepackage{tablefootnote}

\usepackage{graphicx}
\usepackage{tabularray}
\usepackage{xcolor}
\usepackage{ccicons}

\definecolor{gray7}{HTML}{BFBFBF}
\definecolor{gray9}{HTML}{E5E5E5}

\usepackage{tabularray}

\makesavenoteenv{tabular}
\makesavenoteenv{table}

\journal{DFRWS EU 2026}

\begin{document}
\emergencystretch 3em

\begin{frontmatter}




\title{Plug to Place: Indoor Multimedia Geolocation from Electrical Sockets for Digital Investigation}

\renewcommand{\theaffn}{\arabic{affn}}


\author[label1]{Kanwal Aftab}
\author[label2]{Graham Adams}
\author[label1]{Mark Scanlon}

\affiliation[label1]{organization={Forensics and Security Research Group, School of Computer Science},
            addressline={University College Dublin},
            city={Belfield},
            state={Dublin 4},
            country={Ireland}}

\affiliation[label2]{organization={School of Engineering},
            addressline={Case Western Reserve University},
            city={Cleveland},
             state={Ohio},
            country={United States}}






    

\begin{abstract}

Computer vision is a rapidly evolving field, giving rise to powerful new tools and techniques in digital forensic investigation, and shows great promise for novel digital forensic applications. One such application, indoor multimedia geolocation, has the potential to become a crucial aid for law enforcement in the fight against human trafficking, child exploitation, and other serious crimes. While outdoor multimedia geolocation has been widely explored, its indoor counterpart remains underdeveloped due to challenges such as similar room layouts, frequent renovations, visual ambiguity, indoor lighting variability, unreliable GPS signals, and limited datasets in sensitive domains.

This paper introduces a pipeline that uses electric sockets as consistent indoor markers for geolocation, since plug socket types are standardised by country or region. The three-stage deep learning pipeline detects plug sockets (YOLOv11, mAP@0.5 = 0.843), classifies them into one of 12 plug socket types (Xception, accuracy = 0.912), and maps the detected socket types to countries (accuracy = 0.96 at >90\% threshold confidence). To address data scarcity, two dedicated datasets were created: socket detection dataset of 2,328 annotated images expanded to 4,072 through augmentation, and a classification dataset of 3,187 images across 12 plug socket classes. The pipeline was evaluated on the Hotels-50K dataset, focusing on the TraffickCam subset of crowd-sourced hotel images, which capture real-world conditions such as poor lighting and amateur angles. This dataset provides a more realistic evaluation than using professional, well-lit, often wide-angle images from travel websites. This framework demonstrates a practical step toward real-world digital forensic applications. The code, trained models, and the data for this paper are available open source. 
\end{abstract}




\end{frontmatter}












\section{Introduction}
\label{intro}

Human trafficking is a severe global crime affecting millions across all ages, genders, and backgrounds, causing deep personal, community, and societal harm~\cite{walby2025improving}. It entails the illegal trade of people through deception, violence, or exploitation, resulting in forced labour, sexual abuse, and organ trafficking~\cite{dimas2022operations}. Given its severe and long-lasting impact, the fight against human trafficking is explicitly prioritised under 3 of the United Nations Sustainable Development Goals (SDGs)~\cite{unodc}. In addition, child sexual exploitation material (CSEM) investigation is one of the most common case types encountered in digital forensics laboratories within law enforcement agencies worldwide~\cite{HARGREAVES2024DFPulse}.

\begin{figure*}[h]
  \centering
  \includegraphics[trim={0cm 0cm 1cm 0cm},clip,width=\textwidth]{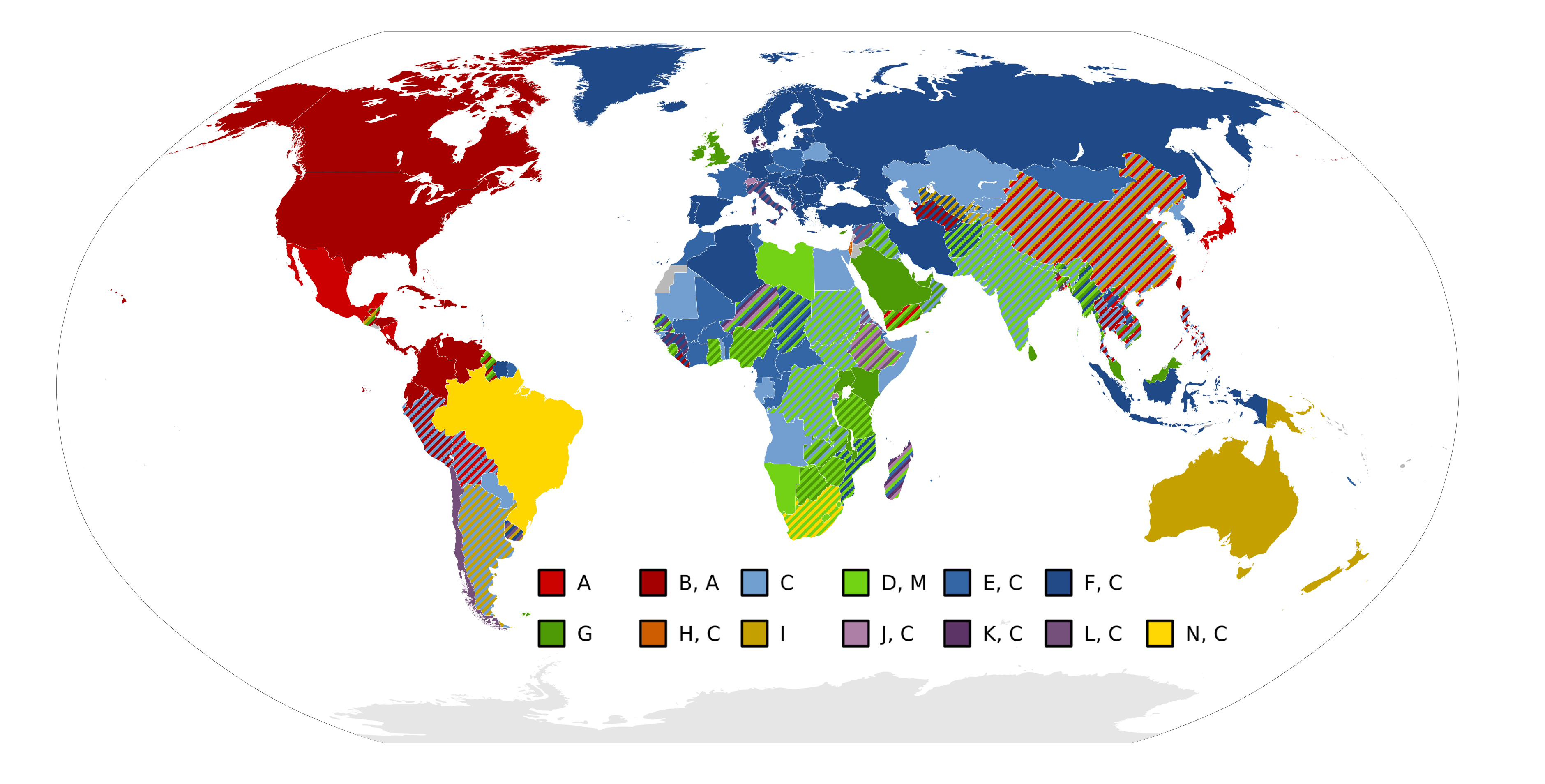}
  \caption{Worldwide plug type distribution map \ccPublicDomain}
  
  \label{fig:plug-distribution}
\end{figure*} 

With rapid technological advancement, the rise of feature-rich smartphones, expanded storage capabilities, widespread internet access, and the growing influence of social media, nearly every facet of modern life has become digital~\cite{Kumar2024}. However, modern technologies are increasingly exploited by offenders to facilitate serious crimes, underscoring the critical importance of digital forensics~\cite{FAHNDRICH2023301617}. As the volume of digital data continues to surge, the analysis and interpretation of digital evidence have become indispensable to modern investigations~\cite{11038544}. The application of artificial intelligence (AI) to digital forensic investigations is still very much in its infancy~\cite{breitinger202410YearReview}, but nonetheless, practitioners have already identified image/media classification as having the most potential for the future use of AI in their investigations~\cite{HARGREAVES2024DFPulse}. In parallel, hotel recognition has become a common need for human trafficking investigations, as hotels are frequently used as intermediary stopover locations during the trafficking of victims~\cite{kamath20212021hotelidcombathuman}. Online human trafficking advertisements or the interception of organised crime gang's electronic communications are often the starting points for many human trafficking investigations. Identifying hotel rooms from these photographs is an extremely arduous task. Indeed, police agencies often resort to crowdsourcing their locations. For example, hotel rooms often feature in the Trace an Object projects run by Europol\footnote{\url{https://www.europol.europa.eu/stopchildabuse}} or the Australian Centre to Counter Child Exploitation\footnote{\url{https://www.accce.gov.au/what-we-do/trace-an-object}}, asking for the general public's help in identifying the hotels during investigations of cases involving child sexual exploitation material. In the DFPulse 2024 survey~\cite{HARGREAVES2024DFPulse}, digital forensic practitioners identified image/media classification and CSAM investigation as two of the main areas where artificial intelligence has the potential to assist in their future cases.

In terms of automated geolocation, indoor geolocation specifically faces significant challenges. GPS is unreliable indoors. Alternative indoor sensors are relatively costly and have limitations in range and accuracy~\cite{HROMADOVA2021882}, and image metadata is frequently stripped during online or instant messaging sharing -- making it difficult to trace an image’s origin~\cite{bamigbade2024}. While outdoor environments typically provide clear geolocation cues, such as landmarks and infrastructure~\cite{ZHANG2021107760, brejcha2017state, luo2011geotagging}, indoor settings are far more complex. Similar layouts, recurring furniture, and inconsistent lighting make it challenging to reliably differentiate between locations~\cite{10440661}. Amid these challenges, it is therefore crucial to identify consistent and distinctive cues for indoor geolocation. One such feature of indoor environments is the presence of electric sockets, which can serve as distinctive and geographically informative visual markers. Each country adheres to standardised socket designs governed by national or regional electrical regulations~\cite{IEC_TR_60083}.
Plug and socket designs vary considerably in shape, grounding, voltage, and frequency. This study focusses solely on their visual characteristics~\cite{worldstandards_plugs}. 
Using Computer Vision (CV), detecting and classifying these plug sockets can provide a reliable cue for narrowing down the search space, as illustrated in Fig.~\ref{fig:plug-distribution}.

\subsection{Contribution of this Work}
This paper makes the following key contributions:
\begin{itemize}

    \item Dataset Creation and Release. Two novel CV datasets have been curated and publicly shared: (i) a socket detection dataset for identifying sockets in indoor room images, and (ii) a socket classification dataset containing 12 socket categories. These datasets provide valuable benchmarks for advancing research in fine-grained indoor object recognition.
    \item Algorithmic Pipeline for Detection and Classification. A comparative study of state-of-the-art detection models and Convolutional Neural Network (CNN) classifiers has been conducted. Based on this analysis, a pipeline was designed to select the most effective combination of detection and classification models for socket recognition.
    \item Evaluation of the Proposed Approach on Real-World Data. The approach is evaluated on the Hotels-50K dataset, specifically the TraffickCam subset, demonstrating its practical utility for law enforcement by narrowing down search spaces in real-world human trafficking investigations. Furthermore, this work aims to lay the foundation for a universal socket detector, enabling broader applications beyond the law enforcement/investigative domains.
\end{itemize}

All code, trained models, and data created as part of this paper are made available open source and can be accessed from \url{https://github.com/markscanlonucd/plugtoplace}.

\section{Related Work}
\label{RelatedWork}

To address the challenge of indoor multimedia geolocation, specifically hotel recognition, researchers have employed a variety of techniques, ranging from hand-crafted feature extraction, image classification to neural networks for automated feature extraction, as well as more advanced approaches such as object-based similarity, image embeddings, and semantic scene understanding. Regardless of the methodology employed, the fundamental building block remains CV. 
Broadly, researchers approach multimedia geolocation in two ways. When a large and representative reference dataset is available, the task is formulated as a Content based Image Retrieval (CBIR) problem; in the absence of such a dataset, geolocation instead relies on universal visual cues, i.e., landmarks or sun angle in outdoor environments, to infer location directly from the image itself~\cite{4587784, 6298404}.

In this context, high-quality data remains essential. However, obtaining such data can be particularly challenging in sensitive scenarios, such as human trafficking investigations. Recognising these ethical and operational challenges, the European Union formalised AI regulations through the AI Act~\cite{5df9814463fe45d4aeb516f3ed946ead, eu_ai_act_2024}, aiming to maximise the benefits of AI while enforcing rigorous ethical and safety standards. 

In the context of data collection and processing, the creation of specialised datasets has been significant. The Hotels-50K dataset~\cite{10.1609/aaai.v33i01.3301726} contains over one million labelled images from 50,000 hotels worldwide, sourced from both travel websites and the TraffickCam mobile application, and is designed specifically for hotel recognition research. The Hotel-ID dataset~\cite{kamath20212021hotelidcombathuman} provides a similarly large-scale resource for the same domain. Both datasets serve as benchmarks for evaluating recognition approaches. 

Deep learning models are often black-box systems, and their complex structures make it difficult to interpret how visual inputs lead to an output, obscuring the specific cues that drive similarity judgments. 
To address this, \citet{Black_2022_WACV} introduced a paired image similarity visualisation technique tailored for Transformer architectures. Their method enables the inspection of attention maps to reveal which regions contribute most to a given similarity score. When comparing ResNet and Vision Transformer (ViT) on datasets such as Hotels-50K, Clean Google LandmarksV2, and Stanford Online Products, the study found that ViT’s attention focused on meaningful cues, such as bed covers, runners, and tiled shower walls. These results emphasise the need for explainable embeddings in visual retrieval. 

Similarly, \citet{wazzan2024context} explored how context affects object matching, finding that a moderate amount of context improves annotation efficiency and retrieval accuracy, while excessive context complicates recognition in ambiguous scenes.

Beyond these pixel-level and embedding-based methods, some research focussed on object-centric retrieval. This approach represents images as groups of distinct objects rather than just overall textures, aiming to bridge the gap between simple image features and what people actually see. For instance, \citet{kim2003central} introduced a method to detect and segment the main object in an image, as subjects are often centred in photos. The authors used relevance feedback on these segments to improve retrieval, though their reliance on colour made it less effective in busy scenes. Later, \citet{pradhan2016prominent} used visual saliency maps to identify object regions and create feature vectors for retrieval. This method worked well for images with a clear main object but was less effective for complex scenes. 

Furthermore, \citet{10440661} argued that hotel recognition cannot be treated as a standard image-classification task due to the extremely large number of hotel classes. To address this, the authors proposed an object-centric ensemble approach in which separate models are trained for different object types. The system employed a Hybrid Vision Transformer combining a ResNet-26 backbone for feature extraction with a ViT classifier across 3,110 classes. This hybrid design delivered superior performance, achieving nearly 80 \% top-1 accuracy on the Hotel-ID 2022 dataset, compared with only 60 \% using a conventional full-image method. This shows the value of pinpointing distinctive features, such as furniture, fixtures, or wall patterns. Additionally, semi-automated labelling pipelines~\cite{wazzan2024context} demonstrate that using less context speeds up labelling without hurting performance, suggesting that focusing on main objects and a moderate amount of context best balances interpretability and efficiency.

Besides these semantic methods, feature-level CBIR focusses on capturing broad visual qualities such as colour, texture, and layout~\cite{10.1007/978-3-030-58565-5_43}. Among these attributes, colour is a dominant and widely used image characteristic, as it is both robust, computationally efficient, and independent of orientation and image resolution, making it highly versatile.~\cite{9945709, 9497165}. While RGB is common, colour spaces like CIELAB, Munsell, and fuzzy-based models often yield more meaningful results~\cite{9945709}. Using a mix of colour spaces, such as RGB, YCbCr, and Lab*, can also boost precision~\cite{sangeetha2022enhanced}. Specific to hotel room identification, \citet{herrmann2024colourbasedgeolocation} evaluated CBIR systems using colour features on Hotels-50K, achieving over 95\% Top-50 accuracy with just two descriptors, thus supporting faster and more reliable investigative workflows. More recently, \citet{bamigbade2025improving} combined major colour palettes and simple histograms with deep metric learning and classification to raise top-20 retrieval accuracy by 17\%. These findings show that blending handcrafted features with deep embeddings can improve both clarity and results.

Even with sophisticated methods, systematic reviews expose key gaps in using CV for social good. \citet{dimas2022operations} note that Operations Research and Analytics efforts for anti-human trafficking mostly target sex trafficking and prosecution, with less attention to labour trafficking, prevention, or victim protection. Similarly, \citet{bamigbade2024} highlighted the benefits of using CV based geolocation in investigations, calling for more data types and clearer deep learning models to extract useful information.

Overall, these studies show a shift in hotel recognition research: from data-driven deep metric learning to more comprehensible, object-focused, and feature-blended CBIR systems. This change not only improves technical performance but also supports the growing demands for ethical, transparent, and socially responsible AI in sensitive investigations.

\begin{figure*}[h]
  \centering
  \includegraphics[width=0.9\textwidth]{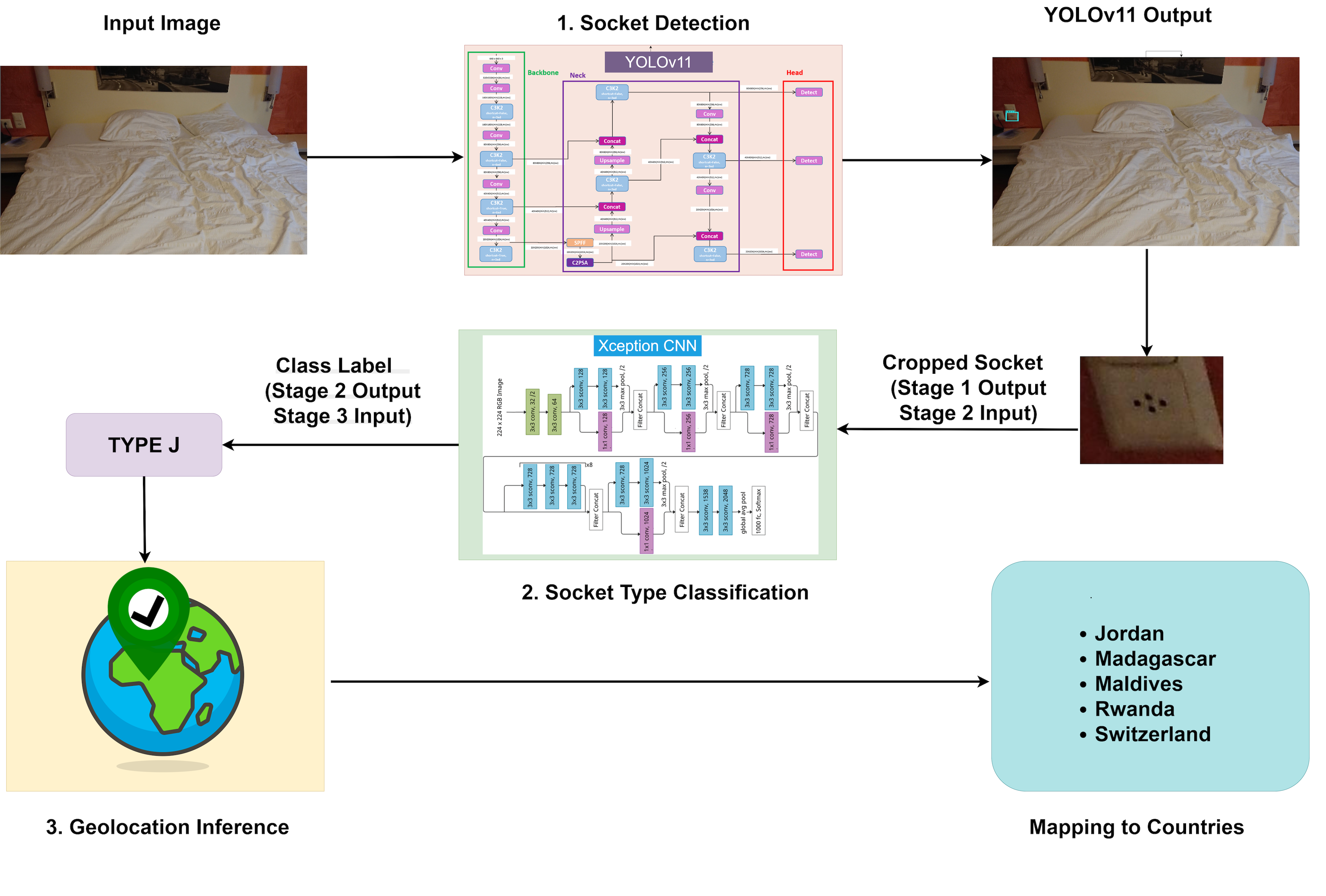}
  \caption[Architecture of the proposed three-stage system]{Architecture of the proposed three-stage pipeline: (1) Socket detection, (2) Socket type classification, and (3) Geolocation.}
  \label{fig:pipeline}
\end{figure*}

\section{Experimental Setup and Analysis}
\label{Experiment}

The CV methodology proposed as part of this paper is a three-tiered process, as detailed in Fig.~\ref{fig:pipeline}. In the first stage, YOLO object detection identifies and localises electric sockets in an image, generating cropped regions of interest (ROI). In the second stage, a CNN classifies each cropped socket ROI by type. Finally, in the third stage, the detected socket type is mapped to potential countries, which narrows the list of likely locations and supports law enforcement investigations. Sections 4 to 6 provide a detailed discussion of each stage, including dataset preparation for every stage, the methodologies used, and a comparative analysis of different algorithms along with their results.

\section{Stage 1: Socket Detection}

Stage 1 of the pipeline focuses on detecting electric sockets in image to infer indoor geographical location. Electric sockets are consistent and recognisable markers, as each country or region uses specific socket types defined by distinct pin configurations. These features are visually distinctive in indoor environments such as hotel rooms. A CV-based object detection model is applied to automatically identify socket instances and their positions within an image. The detected socket type is then used to constrain the possible geographical location, supporting subsequent investigative analysis.

\subsection{Dataset Preparation} 
High-quality and diverse datasets are critical in CV, as both the quantity and quality of training data directly affect model accuracy and generalisation~\cite{joshi2024data}. For this study, a total of 2,328 socket images were compiled, of which 1,525 were obtained from publicly available Roboflow~\cite{roboflow} socket datasets licenced under CC 4.0, while the remaining images were cropped from hotel room scenes in the Hotels-50K dataset. This original dataset (Dataset A) was partitioned into training (70\%, 1,629 images), validation (20\%, 455 images), and test (10\%, 244 images) subsets. All images were annotated with bounding boxes using the Roboflow platform and manually classified into two categories: class 0 (NA) for switchboards and class 1 for sockets. Although the initial focus was exclusively on sockets, the frequent misclassification of visually similar switchboard buttons necessitated the inclusion of the NA class to better differentiate non-target elements from actual sockets. This confusion was heavily compounded by the small sizes of the objects, their low resolution in cropped images, and the similar colouring and high concentration of buttons and sockets on the switchboards.

To enhance dataset diversity and improve model generalisation, data augmentation was applied exclusively to the training set, doubling its size from 1,629 to 3,258 images. Augmentation techniques included random cropping (0–20\%), rotation (–15° to +15°), grayscale conversion (15\% of images), hue adjustment (–24° to +24°), and brightness adjustment (–19\% to +19\%). After additional images were added, the dataset resulted in training (80\%, 3,258 images), validation (10\%, 409 images), and test (10\%, 407 images) subsets, for a total of 4,074 images. The validation and test sets were not augmented; however, an additional 163 images cropped from the Hotels-50K dataset were added to ensure an unbiased evaluation of model performance.

\begin{table*}[ht]
\centering
\caption{Test results for all classes across models. Bold indicates the best value per column.}
\begin{tabular}{l c c c c c c}
\hline
Version & Type & Epochs & Precision (P) & Recall (R) & mAP@0.5 & mAP@0.5:0.95 \\
\hline
YOLOv8 & s & 180 & 0.843 & 0.709 & 0.793 & 0.529 \\
YOLOv8 & s & 250 & 0.834 & 0.709 & 0.790 & 0.524 \\
YOLOv8 & m & set=300 (early stopping at 233) & \textbf{0.883} & 0.750 & 0.809 &  \textbf{0.540} \\
YOLOv11 & s & 180 & 0.847 & 0.734 & 0.804 &  \textbf{0.540} \\
YOLOv11 & m & 250 & 0.846 &  \textbf{0.791} & \textbf{0.832} & 0.539 \\
YOLOv11 & l & set=250 (early stopping at 234) & 0.851 &  0.764 &  0.806 & 0.534\\
YOLOv12 & s & 180 & 0.720 & 0.650 & 0.748 & 0.483 \\
\hline
\end{tabular}
\label{tab:test_all_classes}
\end{table*}

\subsection{Methodology}
Recent advances in CV have positioned object detection as a fundamental task, achieving performance that, in some cases, rivals or even surpasses human capabilities \cite{Borji_2014_CVPR}. A wide range of state-of-the-art algorithms are available for object detection, and the choice of an appropriate model for custom datasets depends heavily on the specific requirements of the application.
Object detection algorithms are broadly classified as two-stage or single-stage detectors. Two-stage models are more accurate but slower, while single-stage models offer faster inference with only a slight accuracy trade-off~\cite{MOHAMMED2025100322}. To handle large volumes of images efficiently, speed was prioritised, rendering single-stage object detectors a more practical option. The YOLO family was chosen for its combination of high speed and competitive accuracy \cite{wang2023comprehensive}, which is suitable for socket detection as a first-stage step before potential refinement. All YOLO models used in this study are initialised with pre-trained weights and subsequently fine-tuned for socket detection, allowing reliable performance with limited task-specific training data.

Although newer YOLO versions introduce architectural improvements, the latest release is not always stable \cite{jegham2024yolo}. A comparative analysis was then conducted. The experimental outcomes, as outlined in Table~\ref{tab:test_all_classes}, demonstrate the performance of each model across diverse metrics, starting with YOLOv8 as a benchmark, followed by YOLOv11 and YOLOv12. Each release provides multiple variants (nano, small, medium, large). This study predominantly focused on small and medium models, and in one case, the large model was also considered to balance computational cost with accuracy. Hyperparameter tuning, particularly the number of training epochs, was evaluated to minimise underfitting and overfitting, and the effect of data augmentation on model performance was also assessed.

\subsection{Performance matrix}
Mean Average Precision (mAP) is a standard metric in object detection that summarises how accurately a model detects and localises objects. It is based on precision and recall, using Intersection over Union (IoU) to measure the overlap between predicted and ground truth bounding boxes. mAP@0.5 computes the average precision at an IoU threshold of 0.5, meaning a detection is considered correct if at least 50 \% of the predicted box overlaps the ground truth, providing a measure of overall detection success. mAP@0.5:0.95 averages performance across multiple IoU thresholds from 0.5 to 0.95, evaluating the model’s ability to both detect objects and precisely localise them. High mAP@0.5:0.95 indicates strong localisation accuracy, while mAP@0.5 primarily reflects general detection performance. These metrics together provide a robust assessment of the model’s effectiveness in identifying and locating objects in images.

\begin{table*}[h]
\centering
\caption{Comparison of YOLOv11m variant with data augmentation and k fold cross validation on and Dataset B. Bold indicates the best value per dataset.}
\begin{tabular}{l l c c c c}
\hline
Dataset & Setting & Precision (P) & Recall (R) & mAP@0.5 & mAP@0.5:0.95 \\
\hline

Test (Dataset B)   & Yolov11m without Aug  & 0.805  & 0.748  & 0.792  & 0.498  \\
                   & Yolov11m with Aug & 0.783 & 0.765 & 0.766 & 0.485 \\
K fold cross validation & Yolov11s with Aug  & \textbf{0.8675} & \textbf{0.7990} & \textbf{ 0.8427} & \textbf{ 0.5771} \\

\hline
\end{tabular}
\label{tab:yolov11_aug_comparison}
\end{table*}

\subsection{Comparative Analysis and Results}
This study evaluated the performance of various YOLO models for socket detection using Dataset A, the original dataset, in the first phase, and subsequently assessed the effect of data augmentation in the second phase on the augmented data, Dataset B. All models were trained with a batch size of 16 and an input resolution of 640 pixels. The optimiser was set to AdamW, which automatically tuned hyperparameters and overrode the default learning rate and momentum values, resulting in an effective learning rate of 0.001667 with momentum fixed at 0.9.

The experiment began with YOLOv8, starting with the small variant (YOLOv8s) to assess the effect of increasing the number of training epochs from 180 to 250. The focus then shifted to the medium variant (YOLOv8m) to evaluate the trade-off between model capacity and performance. Although the number of epochs was initially set to 300, training was stopped early at 233 epochs due to the early stopping mechanism designed to prevent overfitting.

The analysis then progressed to YOLOv11, beginning with the small variant (YOLOv11s) trained for 180 epochs, followed by the medium variant (YOLOv11m) trained for 250 epochs. For the large variant (YOLOv11l), training was scheduled for 250 epochs but was stopped early at 234 epochs due to early stopping. Finally, YOLOv12 was evaluated; however, it achieved comparatively lower performance metrics.

Specifically, YOLOv12 underperformed, achieving mAP@0.5 of 0.748 and mAP@0.5:0.95 of 0.483, with a precision of 0.720 and a recall of 0.650. These results indicate weaker socket detection performance, highlighting that cutting-edge models do not always outperform more established versions such as YOLOv8 and YOLOv11. Newly released models are often unstable and improve over time; therefore, it is generally advisable to allow them to mature before deployment in CV applications~\cite{jiang2025odverse33}.

Validation accuracy should not be solely relied upon, as it may indicate overfitting on the training data and not necessarily reflect performance on unseen test data~\cite{santos2022avoiding}. Therefore, test accuracy is considered more decisive than validation accuracy. On the test set, YOLOv8m (early stopping at 233 epochs) achieves the highest precision (0.883), while both YOLOv8m and YOLOv11s (180 epochs) achieve the highest mAP@0.5:0.95 (0.540). This indicates that these models are highly accurate in correctly detecting sockets, though their recall varies (0.750 for YOLOv8m, 0.734 for YOLOv11s).
In contrast, YOLOv11m (250 epochs) attains the highest recall (0.791) and the highest mAP@0.5 (0.832), indicating strong overall detection coverage and good localisation. 

To further assess the impact of data augmentation, YOLOv11m was trained on Dataset B with augmentation, while YOLOv11s was trained using both augmentation and 5-fold cross-validation. Table~\ref{tab:yolov11_aug_comparison} summarises the results. Adding only augmentation slightly altered performance, with precision decreasing from 0.846 to 0.783, recall increasing from 0.748 to 0.765, mAP@0.5 decreasing from 0.832 to 0.766, and mAP@0.5:0.95 decreasing from 0.539 to 0.485. These results suggest that augmentation improves recall at a modest cost to precision and overall localisation accuracy.

\begin{figure*}[ht]
    \centering
    \begin{tabular}{@{}cccc@{}} 
        \subfloat[]{\includegraphics[trim={0 4cm 0 0},clip,width=0.45\textwidth]{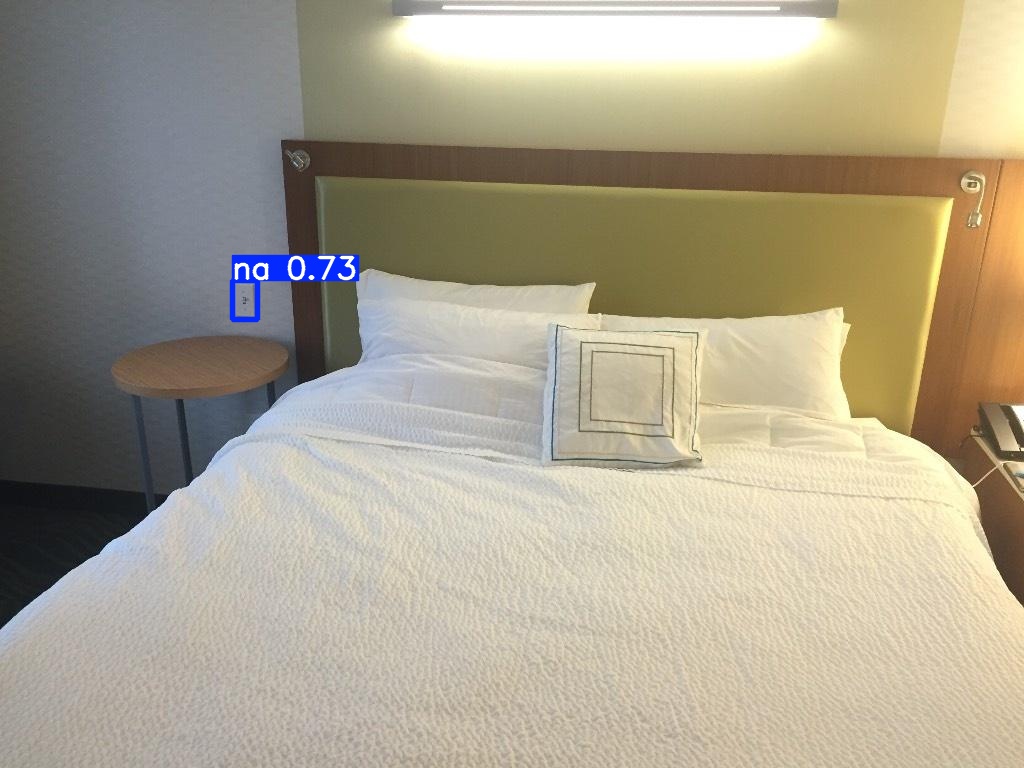}} &
        \subfloat[]{\includegraphics[width=0.45\textwidth]{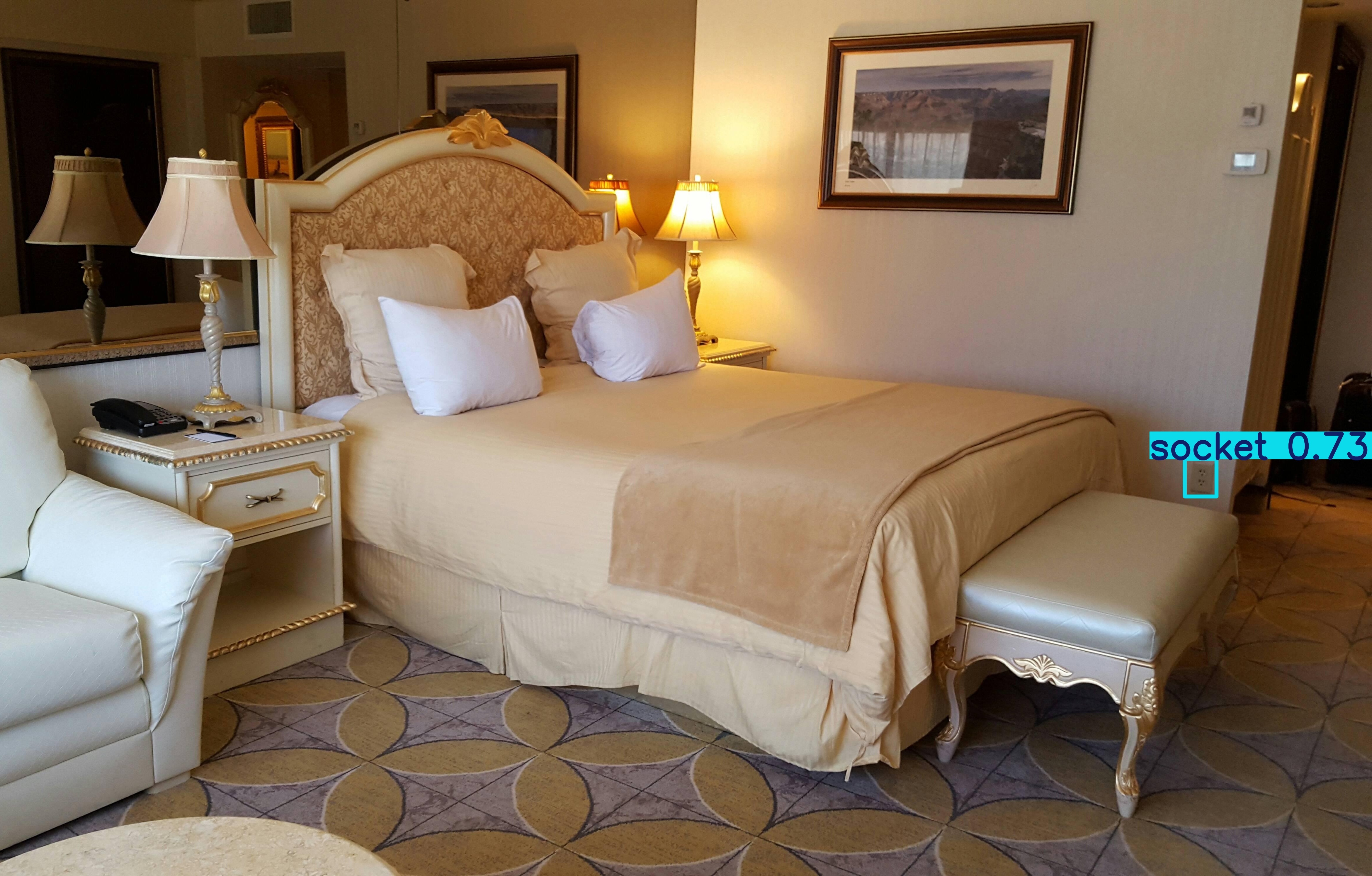}} \\
    \end{tabular}
    \caption{YOLOv11m detection results on room images (a–b), showing bounding boxes for socket classes}
    \label{fig:yolo_roomsamples}
\end{figure*}

\begin{figure}[ht]
    \centering
    \begin{tabular}{@{}cccc@{}} 
        \subfloat[]{\includegraphics[width=4.2cm, height=5cm]{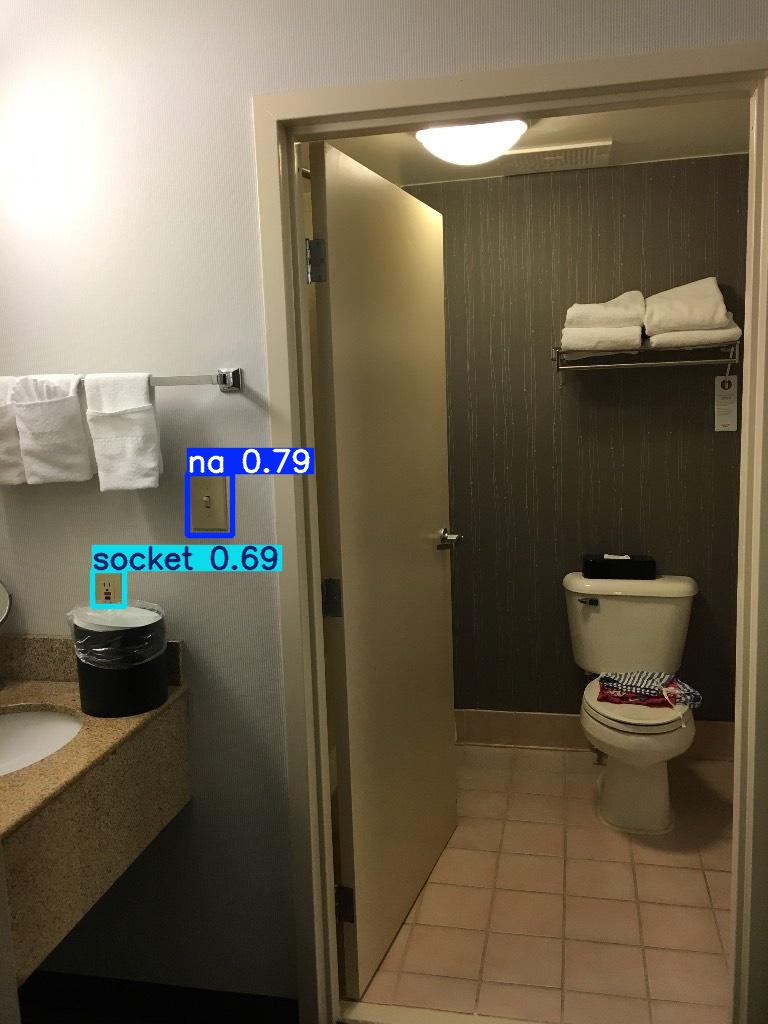}} &
        \subfloat[]{\includegraphics[width=4.2 cm, height=5cm]{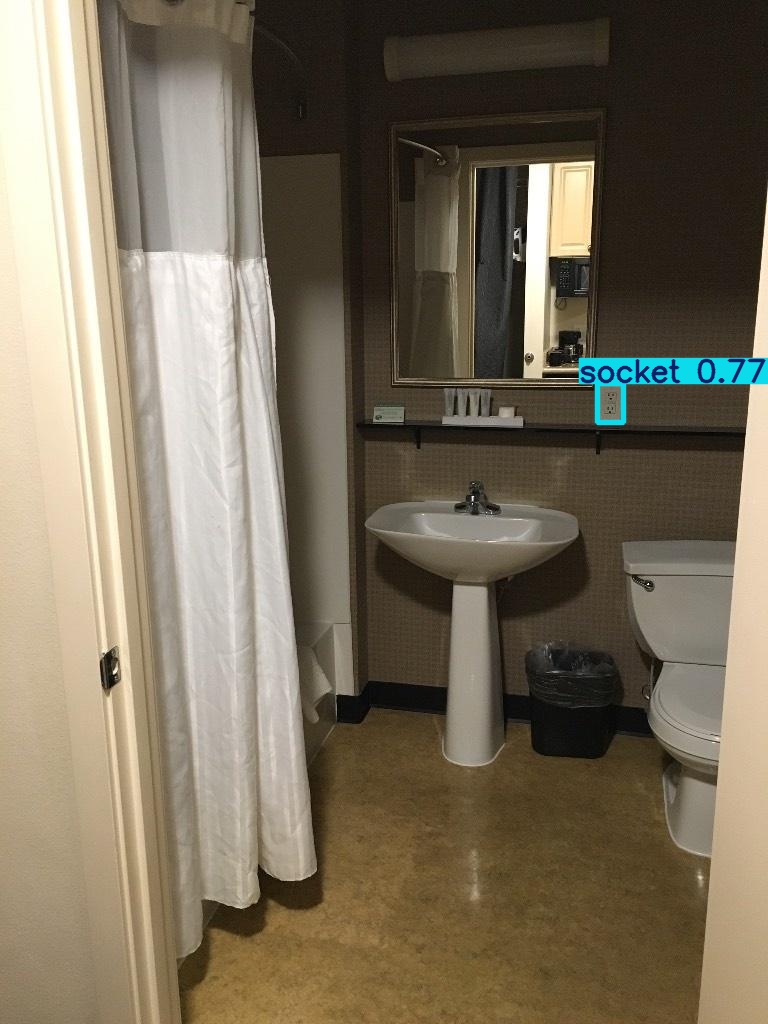}} \\
    \end{tabular}
    \caption{YOLOv11m detection results on bathroom images (a–b), showing bounding boxes for socket classes}
    \label{fig:yolobath_samples}
\end{figure}

In contrast, combining K-fold cross-validation with augmentation further enhances model robustness, reducing variance between folds and providing more reliable generalisation to unseen images. Specifically, average precision increased from 0.847 to 0.872, and recall increased from 0.734 to 0.756, demonstrating that K-fold training effectively mitigates overfitting on smaller datasets and improves overall detection performance. Based on this analysis, YOLOv11s with augmentation and K-fold cross-validation was identified as the best performer in Stage One socket detection. Fig.~\ref{fig:yolo_roomsamples} and Fig.~\ref{fig:yolobath_samples} show the visual socket detection results of this best-performing model.

\section{Stage 2: Socket Type Classification}

\begin{figure*}
    \centering
    \includegraphics[width=0.8\textwidth]{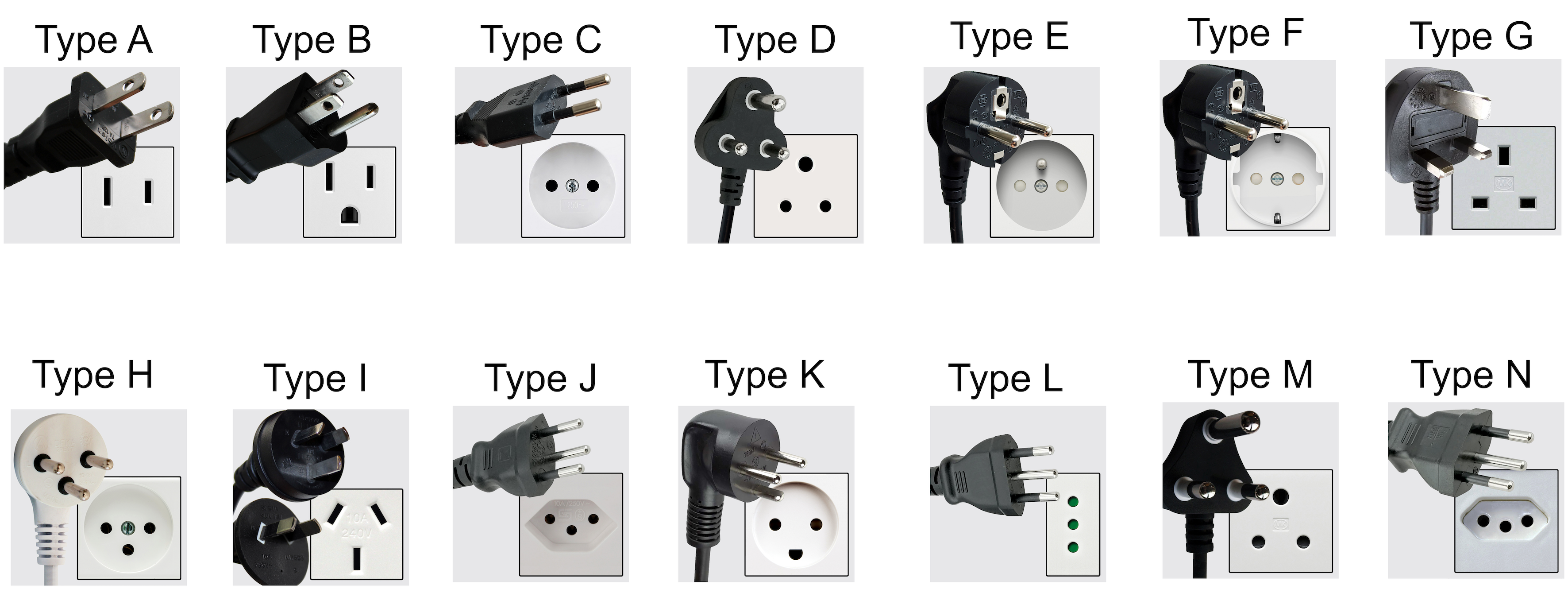}
    \caption{Plug and Socket Types from Type A to Type N~\citep{worldstandards_plugs}}
    \label{fig:14_types}
\end{figure*}

Stage 2 of the pipeline focuses on classifying detected electric sockets into region-specific socket types based on their visual characteristics. Although 14 domestic socket types are internationally recognised, this stage considers only visual distinctions relevant for image-based classification. As illustrated in Fig.~\ref{fig:14_types} and Fig.~\ref{fig:plug-distribution}, socket designs exhibit strong regional variation, enabling geographical constraints to be inferred from socket appearance.

In this experiment, CV is used to classify socket types, since the focus is exclusively on visual characteristics, the study considers 12 classes instead of the full set of 14. Types D and M have been merged into a single class (DM) because, despite differences in pin size, their layouts are visually indistinguishable. Type M closely resembles Type D but features larger pins. Similarly, Types J and N are merged into a single class (JN). Both sockets use three round pins with nearly identical configurations, differing only in the precise offset of the earth pin -- with Type J, it is offset by 5mm, and with Type N, it is offset by 3mm~\cite{IEC_TR_60083}. This small structural difference makes them electrically incompatible, but the variation was deemed too subtle to be reliably distinguished visually.
All other socket types present clear visual differences and are, therefore, treated as separate classes. In addition, a noise class was introduced to exclude non-socket objects, such as light switches, thermostats, or low-quality regions of interest detected by YOLO, thereby further enhancing accuracy. 

\subsection{Dataset Preparation}
The dataset for socket type classification was constructed from two primary sources. The first source was publicly available datasets from Roboflow, consisting of plug socket images licenced under CC 4.0. These images were downloaded, cleaned, and merged into a consistent collection. The second source was the Hotels-50K dataset, from which socket regions were automatically detected, cropped, and assigned to their respective classes. After processing, the final dataset comprised 12 socket type classes, with the number of images per class summarised in Table~\ref{tab:socket_combined}. In total, the dataset contains 3,187 images, which were partitioned into training, validation, and test sets using a 70:15:15 split.
\begin{table}[h!]
\centering
\setlength{\tabcolsep}{6pt} 
\renewcommand{\arraystretch}{1.0} 
\caption{Socket Types with Image Counts and Corresponding Country Usage Count.}
\label{tab:socket_combined}
\begin{tabular}{c c c}
\hline
Socket & Images & Countries \\
\hline
A     & 192 & 46 \\
B     & 302 & 28 \\
C     & 305 & 65 \\
D / M & 300 & 12 / 9 \\
E     & 304 & 24 \\
F     & 303 & 35 \\
G     & 304 & 32 \\
H     & 138 & 1  \\
I     & 268 & 11 \\
J / N & 262 & 5 / 4  \\
K     & 291 & 6  \\
L     & 222 & 9  \\
\hline
Total & 3,187 & - \\
\hline
\end{tabular}
\end{table}

\subsection{Methodology}
A transfer learning approach for multi-class image classification using five state-of-the-art CNN architectures was implemented: VGG16, InceptionV3, Xception, ResNet50, and ResNet101. The dataset, consisting of 3,187 images across 12 classes (socket types), was split into training, validation, and test sets, with 2,224 images for training, 473 images for validation, and 490 images for testing. Images were preprocessed and augmented with normalisation and horizontal flipping to improve generalisation. Each model was initialised with ImageNet pre-trained weights, and the convolutional base was frozen to leverage pre-learned features while training a new classification head for the target classes. Models were trained independently using categorical cross-entropy loss and the Adam optimiser. 

\subsection{Evaluation and Comparative Analysis}

The performance of the model was evaluated using standard metrics, including accuracy, precision, recall, F1-score, and a confusion matrix to provide detailed insights into class-wise predictions. Among the evaluated models, VGG16, InceptionV3, and Xception achieved the highest accuracies. Xception attained the best overall performance with an accuracy of 91.22\%, consistently demonstrating high precision, recall, and F1-scores across all 12 classes, as summarised in Table~\ref{tab:model_performance}. In comparison, VGG16 achieved 82.65\% accuracy, while InceptionV3 reached 89.80\%, highlighting the performance differences among the individual models.
\begin{table}[h!]
\centering
\caption{Performance Summary of Different Models}
\begin{tabular}{lcccc}
\hline
\textbf{Model} & \textbf{Accuracy} & \textbf{Precision} & \textbf{Recall} & \textbf{F1-score} \\
\hline
VGG16        & 0.827 & 0.846 & 0.816 & 0.819 \\
InceptionV3  & 0.898 & 0.907 & 0.900 & 0.901 \\
Xception     & \textbf{0.912} & \textbf{0.914} & \textbf{0.910} & \textbf{0.911} \\
ResNet50     & 0.492 & 0.599 & 0.477 & 0.466 \\
ResNet101    & 0.443 & 0.634 & 0.429 & 0.433 \\
\hline
\label{tab:model_performance}
\end{tabular}
\end{table}

To further validate the robustness of Xception, the impact of K-fold cross-validation was evaluated by modifying the dataset. A new Noise class, consisting of 304 images representing non-socket objects potentially missed by Step 1 YOLO, was added to the original 3,187 images, resulting in a total of 3,495 images across 13 classes (12 socket types + Noise). These images were organised into 13 folders for training and validation in a 5-fold cross-validation setup, while an additional 175 unseen images, spanning all classes, were reserved for testing. This setup allowed us to assess whether cross-validation improves model generalisation.

As summarised in Table~\ref{tab:xception_performance}, the 5-fold cross-validation strategy increased accuracy from 85.4\% to 87.7\%, precision from 87.8\% to 89.4\%, recall from 85.3\% to 88.4\%, and F1-score from 85.5\% to 88.1\%. These results demonstrate that K-fold cross-validation provided a modest but consistent improvement in performance over the single-run Xception model. Note that the test dataset differs from the split of 70:15:15.

\begin{table}[h!]
\centering
\small
\setlength{\tabcolsep}{4pt} 
\caption{Performance Summary of Xception Models}
\begin{tabular}{lcccc}
\hline
\textbf{Model} & \textbf{Accuracy} & \textbf{Precision} & \textbf{Recall} & \textbf{F1} \\
\hline
Xception             & 0.854 & 0.878 & 0.853 & 0.855 \\
Xception (5-Fold CV) & \textbf{0.877} & \textbf{0.894} & \textbf{0.884} & \textbf{0.881} \\
\hline
\end{tabular}
\label{tab:xception_performance}
\end{table}

Therefore, Xception with 5 cross-validation was identified as the best performer in stage two socket type classification.

\section{Stage 3: Inferring Geolocation through Socket Type–Country Mapping}
The final stage of the proposed pipeline focusses on using socket detection results to infer the geolocation of hotel rooms, highlighting the practical applicability of the proposed approach.
\subsection{Test Dataset Preparation}
The test dataset for this experiment was derived from the Hotels-50K TraffickCam dataset, available from its official GitHub repository. To prepare the data, the dataset was obtained by modifying the download script to preserve the original image resolution. This was a crucial step because the target object, electric sockets, is often small within the larger image, and resizing could lead to a loss of detail essential for accurate detection.

The Hotels-50K dataset comprises two subcategories: TraffickCam and travel website images. TraffickCam consists of crowd-sourced photos submitted by travellers worldwide, reflecting real-world conditions. After retaining and downloading the images in their original resolution of 1024 × 768, the dataset provides images in this resolution, with sockets typically appearing in small regions of approximately 130 × 87 pixels.

In contrast, travel website images are professionally captured under ideal lighting and angles, often with colour correction and photo editing. These images are also provided at a lower resolution (350 × 233), making the sockets barely visible. As a result, they were excluded from testing. The TraffickCam subset, by depicting more realistic, non-professionally taken and/or edited photos, is considered to provide a better representation of the conditions encountered in practical investigations and was therefore used exclusively for evaluation.

To establish a ground truth for the experiment, the dataset was processed to associate each image with its corresponding country. This involved a multistep process:
Merging Metadata: the various CSV files from the original dataset were consolidated to create a unified file containing image IDs, hotel IDs, and geographic coordinates (latitude and longitude).
Geolocation: Using the \texttt{geopy.geocoders} library, the geographic coordinates were converted into country names.
Standardisation: To ensure consistency, the country names were standardised using the \texttt{pycountry} library, as the raw geolocation output sometimes returned names in native languages. Illegal characters were removed, and spaces were replaced with underscores to create valid directory names.

The final dataset was restructured into a clean directory, with images organised into subfolders named after their respective countries. This structure, along with a consolidated CSV file containing all relevant metadata, streamlined the subsequent country-specific analysis. The reason images were arranged in folders instead of being directly taken and compared from the CSV file was to facilitate visual inspection, ensuring that the code functioned correctly and that each country was accurately represented with its respective socket type.

\subsection{Data Analysis, Evaluation, and Results}

The model's performance was evaluated by assessing its ability to detect a plug socket in an image and then classify it to determine its corresponding country. 
For each image in which a socket was successfully detected and classified, it was checked whether the predicted plug type and the image's actual country formed a valid pair according to this mapping. A correct match was assigned a score of 1, an incorrect match received a score of $-1$, and the noise class was assigned a neutral score of 0. These scores were used to calculate key performance metrics, such as the confusion matrix, precision, recall, and F1 score. This approach provided a comprehensive assessment of the model’s accuracy. Finally, the results were compiled into a detailed summary report. They were presented with graphical visualisations, such as bar charts, to provide a clear and intuitive interpretation of the model's overall performance.

A total of 44,630 TraffickCam images were processed through the algorithmic pipeline. In the first stage, YOLO detected 3,759 potential sockets. To enhance detection accuracy and eliminate false positives, a second-stage classifier was employed to identify and remove noise. Specifically, instances where non-socket objects (e.g., switchboards) were incorrectly detected as sockets in the first stage were classified as noise. This step identified 1,393 noisy detections, leaving 2,366 valid sockets. These valid detections were subsequently passed to the socket classification stage, where only those with a confidence threshold above 70\% were retained. The results are summarised in Table~\ref{tab:classifier_threshold}. 

\begin{table}[ht]
\centering
\caption{Classifier Threshold Analysis for Socket Detection}
\resizebox{\columnwidth}{!}{%
\begin{tabular}{c c c c c}
\hline
Class Confidence & Correct & Wrong & Total & Accuracy (\%) \\
\hline
$\geq 70\%$ & 1595 & 146 & 1741 & 91.61 \\
$\geq 80\%$ & 1421 & 95 & 1516 & 93.73 \\
$\geq 90\%$ & 1167 & 45 & 1212 & 96.29 \\
\hline
\end{tabular}
}
\label{tab:classifier_threshold}
\end{table}

\begin{figure}[h]
  \centering
  \includegraphics[width=0.4\textwidth]{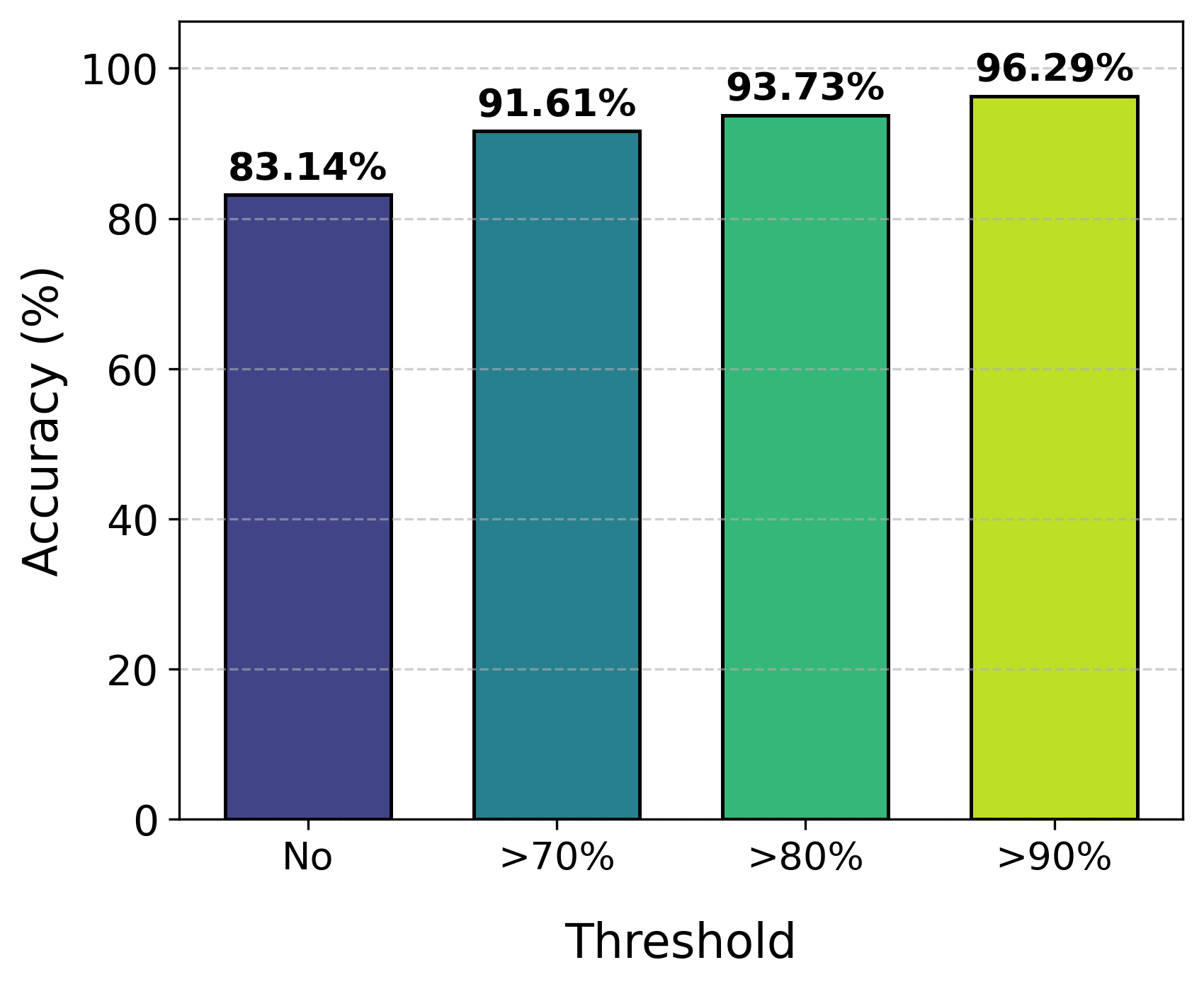}
  \caption{Country prediction accuracy by varying socket type confidence threshold values}
  
  \label{fig:plug-analysis}
\end{figure}

When considering different socket classification confidence thresholds, the performance varies. The confidence threshold represents the model’s predicted probability that a socket belongs to the predicted class. Without applying any socket confidence threshold, 1,967 predictions were correct, and 399 were incorrect, resulting in an accuracy of 83.08\%. At a threshold above 70\%, 1,595 predictions were correct, and 146 were incorrect, resulting in an accuracy of 91.61\%. Increasing the threshold above 80\% reduced the number of correct detections to 1,421, while incorrect detections decreased to 95, yielding an improved accuracy of 93.73\%. At the highest threshold of above 90\%, correct detections further decreased to 1,167, with only 45 incorrect predictions, resulting in the highest accuracy of 96.29\%. These results illustrate the trade-off between confidence and accuracy: lower thresholds capture more sockets but result in more false positives, whereas higher thresholds reduce errors at the cost of missing some detections, as shown in Fig.~\ref{fig:plug-analysis}.

\section{Discussion}
\label{Discussion}

The proposed pipeline has been developed with generalisability in mind, aiming to detect electric sockets in images beyond the Hotels-50K dataset. As a universal visual cue, sockets have the potential to support geolocation in any indoor image. However, socket detection is inherently challenging due to their small size and the frequent low resolution of imagery. Despite these challenges, the method performs strongly when sockets are present, with geolocation inference achieving 96.29 \% accuracy. The reported precision of ~0.85 demonstrates effective detection, but only around 10 \% of TraffickCam images contain visible sockets, which explains why applying high confidence thresholds reduces usable images to about 2 \% of the dataset.
Misidentification of countries is primarily due to the YOLO-based detection stage identifying non-socket objects, such as switchboards or other sources of visual noise, as potential sockets. In stage two, these detections are assigned the NOISE class, resulting in a incorrect socket score. Contributing factors include low image resolution, occlusions, and challenging lighting conditions. These issues indicate that most misclassifications are technical limitations of the current detection and classification pipeline, although some inherent limitations remain when atypical or uncommon socket types are present. Improvements in detection accuracy, higher-resolution imagery, and enhanced classification models could reduce such errors.

\section{Conclusion}
\label{Conclusion}

This paper presents a universal pipeline for socket detection and classification, motivated by its potential to enhance geolocation capabilities in investigative contexts where conventional cues, such as metadata, outdoor landmarks, and sensor data, are unavailable. By curating two new datasets, benchmarking multiple detection and classification models, and evaluating performance on the Hotels-50K TraffickCam dataset, this study demonstrates both the feasibility and the challenges associated with socket-based indoor geolocation. Despite challenges such as small object size and low-resolution imagery, the results demonstrate strong detection accuracy and validate the concept’s practical potential, enabling investigators to automate geolocation across large volumes of digital evidence and convert it into actionable intelligence.

Beyond technical performance, this work contributes to the emerging field of AI-driven multimedia forensics, where visual scene elements are utilised to support digital investigations. Socket detection offers a unique, region-specific forensic cue that can aid law enforcement agencies in narrowing search regions, corroborating other forms of evidence, and identifying the possible origins of illicit or trafficking-related imagery. The proposed framework, therefore, represents an important step towards scalable, context-aware forensic tools that bridge the gap between CV and real-world investigative practice.

Future work will focus on grouping visually similar scenes based on distinctive feature representations, enabling indirect geolocation inference and the propagation of cues from one image across an entire image set. Such clustering can reveal patterns in large datasets and link unknown or unlabelled images to geographically known locations. Further integration of multimodal cues, including architectural styles, interior designs, or visible textual information, could strengthen geolocation when electrical fixtures are unavailable. Human inspection of representative subsets could also help verify socket presence and detection accuracy, refine confidence thresholds, and assess the pipeline’s applicability to diverse, real-world images. Together, these strategies would make the system more scalable and practical for forensic and investigative applications, bridging the gap between AI-based contextual geolocation and digital investigations. to cluster visually similar scenes and strengthen geolocation inference when sockets are not visible. By combining socket-based detection with broader contextual cues, the framework could evolve into a robust and general-purpose forensic system capable of supporting a wide range of investigative and humanitarian applications.

\bibliographystyle{elsarticle-harv}

\bibliography{sample-base}

@String(ECCV= {Eur. Conf. Comput. Vis.})

@String(AAAI = {AAAI})

@String(ECCV  = {ECCV})

@INPROCEEDINGS{9497165,
  author={Bakair, Ahmed S. and Mohamed, Taha M. and Sadek, Rowayda A.},
  booktitle={2017 27th International Conference on Computer Theory and Applications (ICCTA)}, 
  title="{Enhancing The Capability of Color based CBIR by Voting of Color Histograms}", 
  year={2017},
  volume={},
  number={},
  pages={4-10},
  doi={10.1109/ICCTA43079.2017.9497165}}

@article{walby2025improving,
author={Walby, Sylvia
and Francis, Brian},
title="{Improving the Estimate of Trafficking in Human Beings and Modern Slavery by Integrating Data From ILO/Walk Free/IOM and UNODC}",
journal={Social Indicators Research},
year={2025},
month={Jan},
day={01},
volume={176},
number={2},
pages={669-693},
issn={1573-0921},
doi={10.1007/s11205-024-03474-w},
urlREM={https://doi.org/10.1007/s11205-024-03474-w}
}

@article{dimas2022operations,
    doi = {10.1371/journal.pone.0273708},
    author = {Dimas, Geri L. AND Konrad, Renata A. AND Lee Maass, Kayse AND Trapp, Andrew C.},
    journal = {PLOS ONE},
    publisher = {Public Library of Science},
    title = {Operations research and analytics to combat human trafficking: A systematic review of academic literature},
    year = {2022},
    month = {08},
    volume = {17},
    url = {https://doi.org/10.1371/journal.pone.0273708},
    pages = {1-24},
    number = {8},

}

@misc{unodc,
    author = "{United Nations Office on Drugs and Crime}",
    title = "{Human Trafficking and the SDGs}",
    year = "2025",
    note = "\url{https://www.unodc.org/unodc/human-trafficking/sdgs.html}"
}

@article{ZHANG2021107760,
title = "{Visual place recognition: A survey from deep learning perspective}",
journal = {Pattern Recognition},
volume = {113},
pages = {107760},
year = {2021},
issn = {0031-3203},
doi = {https://doi.org/10.1016/j.patcog.2020.107760},
url = {https://www.sciencedirect.com/science/article/pii/S003132032030563X},
author = {Xiwu Zhang and Lei Wang and Yan Su}
}

@article{brejcha2017state,
author = {Brejcha, Jan and \v{C}ad\'{\i}k, Martin},
title = {State-of-the-art in visual geo-localization},
year = {2017},
issue_date = {August 2017},
publisher = {Springer-Verlag},
address = {Berlin, Heidelberg},
volume = {20},
number = {3},
issn = {1433-7541},
urlREM = {https://doi.org/10.1007/s10044-017-0611-1},
doi = {10.1007/s10044-017-0611-1},
journal = {Pattern Analysis and Applications},
month = aug,
pages = {613–637},
numpages = {25}
}

@article{luo2011geotagging,
author={Luo, Jiebo
and Joshi, Dhiraj
and Yu, Jie
and Gallagher, Andrew},
title={Geotagging in multimedia and computer vision---A survey},
journal={Multimedia Tools and Applications},
year={2011},
month={Jan},
day={01},
volume={51},
number={1},
pages={187-211},
issn={1573-7721},
doi={10.1007/s11042-010-0623-y},
url={https://doi.org/10.1007/s11042-010-0623-y}
}

@article{breitinger202410YearReview,
title = "{DFRWS EU 10-year review and future directions in Digital Forensic Research}",
journal = {Forensic Science International: Digital Investigation},
volume = {48},
pages = {301685},
year = {2024},
note = "{DFRWS EU 2024 - Selected Papers from the 11th Annual Digital Forensics Research Conference Europe}",
issn = {2666-2817},
doi = {https://doi.org/10.1016/j.fsidi.2023.301685},
url = {https://www.sciencedirect.com/science/article/pii/S2666281723002044},
author = {Frank Breitinger and Jan-Niclas Hilgert and Christopher Hargreaves and John Sheppard and Rebekah Overdorf and Mark Scanlon}
}

@misc{bamigbade2024,
      title="{Computer Vision for Multimedia Geolocation in Human Trafficking Investigation: A Systematic Literature Review}", 
      author={Opeyemi Bamigbade and John Sheppard and Mark Scanlon},
      year={2024},
      eprint={2402.15448},
      archivePrefix={arXiv},
      primaryClass={cs.CV},
      url={https://arxiv.org/abs/2402.15448}, 
}

@article{5df9814463fe45d4aeb516f3ed946ead,
title = "{General-purpose AI regulation and the European Union AI Act}",
author = "Gstrein, \{Oskar J.\} and Noman Haleem and Andrej Zwitter",
year = "2024",
month = aug,
day = "1",
doi = "10.14763/2024.3.1790",
language = "English",
volume = "13",
pages = "1--26",
journal = "Internet Policy Review",
issn = "2197-6775",
publisher = "Alexander von Humboldt Institute for Internet and Society",
number = "3",
}

@Inbook{Kumar2024,
author="Kumar, Sachin
and Verma, Ajit Kumar
and Mirza, Amna",
title="Digital Revolution, Artificial Intelligence, and Ethical Challenges",
bookTitle="Digital Transformation, Artificial Intelligence and Society: Opportunities and Challenges",
year="2024",
publisher="Springer Nature Singapore",
address="Singapore",
pages="161--177",
isbn="978-981-97-5656-8",
doi="10.1007/978-981-97-5656-8_11",
urlREM="https://doi.org/10.1007/978-981-97-5656-8_11"
}

@misc{IEC_TR_60083,
  author = {{International Electrotechnical Commission}},
  title = "{Plugs and socket-outlets for domestic and similar general use standardized in member countries of IEC}",
  institution = {{International Electrotechnical Commission}},
  number = {IEC/TR 60083},
  year = {2015},
  edition = {7.0},
  url = {https://webstore.iec.ch/publication/23628},
  note = "{Technical Report}"
}

@misc{eu_ai_act_2024,
  author       = {{European Union}},
  title        = {{Regulation (EU) 2024/1689 of the European Parliament and of the Council of 13 June 2024 laying down harmonised rules on artificial intelligence and amending Regulations (EC) No 300/2008, (EU) No 167/2013, (EU) No 168/2013, (EU) 2018/858, (EU) 2018/1139 and (EU) 2019/2144 and Directives 2014/90/EU, (EU) 2016/797 and (EU) 2020/1828 (Artificial Intelligence Act)}} ,
  year         = {2024},
  howpublished = {Official Journal of the European Union, L2024/1689},
  note         = "{Accessed: 16-12-2025}",
  url          = {https://eur-lex.europa.eu/legal-content/EN/TXT/?uri=CELEX%3A32024R1689}
}

@article{HROMADOVA2021882,
title = {Impact of user orientation on indoor localization based on Wi-Fi},
journal = {Transportation Research Procedia},
volume = {55},
pages = {882-889},
year = {2021},
note = "{14th International Scientific Conference on Sustainable, Modern and Safe Transport}",
issn = {2352-1465},
doi = {https://doi.org/10.1016/j.trpro.2021.07.056},
url = {https://www.sciencedirect.com/science/article/pii/S2352146521004609},
author = {Veronika Hromadová and Juraj Machaj and Peter Brída}
}

@INPROCEEDINGS{11038544,
  author={Khadka, Sujan and Singh, Roshan and Begum, Masrath and Muhammad, Osama and Naeem, Bushra and Chaeikar, Saman Shojae},
  booktitle={2024 IEEE Consumer Life Tech (ICLT)}, 
  title="{A Review on Computer Crimes and Digital Forensics}", 
  year={2024},
  volume={},
  number={},
  pages={1-4},
  doi={10.1109/ICLT63507.2024.11038544}}

@article{FAHNDRICH2023301617,
title = "{Digital forensics and strong AI: A structured literature review}",
journal = {Forensic Science International: Digital Investigation},
volume = {46},
pages = {301617},
year = {2023},
issn = {2666-2817},
doi = {https://doi.org/10.1016/j.fsidi.2023.301617},
url = {https://www.sciencedirect.com/science/article/pii/S2666281723001294},
author = {Johannes Fähndrich and Wilfried Honekamp and Roman Povalej and Heiko Rittelmeier and Silvio Berner and Dirk Labudde}
}

@misc{roboflow,
  author       = {Roboflow},
  title        = "{Roboflow: Computer Vision Tools for Developers and Enterprises}",
  year         = 2025,
  url          = {https://roboflow.com/},
  note         = "{Accessed: 18-09-2025}"
}

@INPROCEEDINGS{10440661,
  author={Bhavanasi, Sai Shreyas and Stylianou, Abby},
  booktitle={2023 IEEE Applied Imagery Pattern Recognition Workshop (AIPR)}, 
  title="{Hotel Recognition Using Object Ensembles}", 
  year={2023},
  volume={},
  number={},
  pages={1-8},
  doi={10.1109/AIPR60534.2023.10440661}}

@inproceedings{10.1007/978-3-030-58565-5_43,
author = {Cao, Bingyi and Araujo, Andr\'{e} and Sim, Jack},
title = {Unifying Deep Local and Global Features for Image Search},
year = {2020},
isbn = {978-3-030-58564-8},
publisher = {Springer-Verlag},
address = {Berlin, Heidelberg},
urlREM = {https://doi.org/10.1007/978-3-030-58565-5_43},
doi = {10.1007/978-3-030-58565-5_43},
booktitle = {Computer Vision – ECCV 2020: 16th European Conference, Glasgow, UK, August 23–28, 2020, Proceedings, Part XX},
pages = {726–743},
numpages = {18},
location = {Glasgow, United Kingdom}
}

@article{sangeetha2022enhanced,
author = {Sangeetha, S. K. B. and Mathivanan, Sandeep Kumar and Pandi, Thanapal and Arivu selvan, K. and Jayagopal, Prabhu and Teshite Dalu, Gemmachis},
title = "{An Enhanced Triadic Color Scheme for Content-Based Image Retrieval}",
journal = {Mathematical Problems in Engineering},
volume = {2022},
number = {1},
pages = {5736630},
doi = {https://doi.org/10.1155/2022/5736630},
url = {https://onlinelibrary.wiley.com/doi/abs/10.1155/2022/5736630},
eprintREM = {https://onlinelibrary.wiley.com/doi/pdf/10.1155/2022/5736630},
year = {2022}
}

@INPROCEEDINGS{9945709,
  author={Shamoi, Pakizat and Sansyzbayev, Daniyar and Abiley, Nurmukhamed},
  booktitle={2022 International Conference on Smart Information Systems and Technologies (SIST)}, 
  title="{Comparative Overview of Color Models for Content-Based Image Retrieval}", 
  year={2022},
  volume={},
  number={},
  pages={1-6},
  doi={10.1109/SIST54437.2022.9945709}}

@InProceedings{10.1007/978-3-030-85099-9_24,
author="Tseytlin, Boris
and Makarov, Ilya",
editor="Rojas, Ignacio
and Joya, Gonzalo
and Catal{\`a}, Andreu",
title="{Hotel Recognition via Latent Image Embeddings}",
booktitle="Advances in Computational Intelligence",
year="2021",
publisher="Springer International Publishing",
address="Cham",
pages="293--305",
isbn="978-3-030-85099-9",
doi="10.1007/978-3-030-85099-9_24"
}

@INPROCEEDINGS{pradhan2016prominent,
  author={Pradhan, Jitesh and Pal, Arup Kumar and Banka, Haider},
  booktitle={2016 Fourth International Conference on Parallel, Distributed and Grid Computing (PDGC)}, 
  title="{A prominent object region detection based approach for CBIR application}", 
  year={2016},
  volume={},
  number={},
  pages={447-452},
  doi={10.1109/PDGC.2016.7913237}}

@inproceedings{kim2003central,
author="Kim, Sungyoung
and Park, Soyoun
and Kim, Minhwan",
editor="Bakker, Erwin M.
and Lew, Michael S.
and Huang, Thomas S.
and Sebe, Nicu
and Zhou, Xiang Sean",
title="Central Object Extraction for Object-Based Image Retrieval",
booktitle="Image and Video Retrieval",
year="2003",
publisher="Springer Berlin Heidelberg",
address="Berlin, Heidelberg",
pages="39--49",
url = {https://link.springer.com/chapter/10.1007/3-540-45113-7_5#citeas},
isbn="978-3-540-45113-6"
}

@article{HARGREAVES2024DFPulse,
title = "{DFPulse: The 2024 digital forensic practitioner survey}",
journal = {Forensic Science International: Digital Investigation},
volume = {51},
pages = {301844},
year = {2024},
issn = {2666-2817},
doi = {https://doi.org/10.1016/j.fsidi.2024.301844},
url = {https://www.sciencedirect.com/science/article/pii/S2666281724001719},
author = {Christopher Hargreaves and Frank Breitinger and Liz Dowthwaite and Helena Webb and Mark Scanlon}
}

@misc{kamath20212021hotelidcombathuman,
  publtype={informal},
  author={Rashmi Kamath and Gregory Rolwes and Samuel Black and Abby Stylianou},
  title="{The 2021 Hotel-ID to Combat Human Trafficking Competition Dataset}",
  year={2021},
  cdate={1609459200000},
  journal={CoRR},
  volume={abs/2106.05746},
  url={https://arxiv.org/abs/2106.05746}
}

@inproceedings{10.1609/aaai.v33i01.3301726,
author = {Stylianou, Abby and Xuan, Hong and Shende, Maya and Brandt, Jonathan and Souvenir, Richard and Pless, Robert},
title = "{Hotels-50K: A Global Hotel Recognition Dataset}",
year = {2019},
isbn = {978-1-57735-809-1},
publisher = {AAAI Press},
pages={726-733},
url = {https://doi.org/10.1609/aaai.v33i01.3301726},
doi = {10.1609/aaai.v33i01.3301726},
booktitle = {Proceedings of the Thirty-Third AAAI Conference on Artificial Intelligence and Thirty-First Innovative Applications of Artificial Intelligence Conference and Ninth AAAI Symposium on Educational Advances in Artificial Intelligence},
articleno = {90},
numpages = {8},
location = {Honolulu, Hawaii, USA},
series = {AAAI'19/IAAI'19/EAAI'19}
}

@inproceedings{wazzan2024context,
author = {Wazzan, Albatool and Ahmad, Imtiaz and Macneil, Stephen and Souvenir, Richard},
title = "{Context or Clutter? Efficiently Matching Objects Across Scenes}",
year = {2024},
isbn = {9798400706196},
publisher = {Association for Computing Machinery},
address = {New York, NY, USA},
urlREM = {https://doi.org/10.1145/3652583.3658090},
doi = {10.1145/3652583.3658090},
booktitle = {Proceedings of the 2024 International Conference on Multimedia Retrieval},
pages = {404–413},
numpages = {10},
location = {Phuket, Thailand},
series = {ICMR '24}
}

@INPROCEEDINGS{Borji_2014_CVPR,
  author={Borji, Ali and Itti, Laurent},
  booktitle={2014 IEEE Conference on Computer Vision and Pattern Recognition}, 
  title="{Human vs. Computer in Scene and Object Recognition}", 
  year={2014},
  volume={},
  number={},
  pages={113-120},
  doi={10.1109/CVPR.2014.22}}

@inproceedings{wang2023comprehensive,
  title="{A comprehensive survey on object detection YOLO}",
  author={Wang, Xiangheng and Li, Hengyi and Yue, Xuebin and Meng, Lin},
  booktitle={The 5th International Symposium on Advanced Technologies and Applications in the Internet of Things (ATAIT 2023)},
  pages={77--89},
  year={2023}
}

@misc{worldstandards_plugs,
  author       = {Conrad H. McGregor},
  title        = {{Plug \& socket types around the world}},
  year         = {2025},
  url          = {https://www.worldstandards.eu/electricity/plugs-and-sockets/},
  note         = "{Accessed: 18-09-2025}"
}

@article{santos2022avoiding,
author = {Santos, Claudio Filipi Gon\c{c}alves Dos and Papa, Jo\~{a}o Paulo},
title = "{Avoiding Overfitting: A Survey on Regularization Methods for Convolutional Neural Networks}",
year = {2022},
issue_date = {January 2022},
publisher = {Association for Computing Machinery},
address = {New York, NY, USA},
volume = {54},
number = {10s},
issn = {0360-0300},
urlREM = {https://doi.org/10.1145/3510413},
doi = {10.1145/3510413},
journal = {ACM Computing Surveys},
month = sep,
articleno = {213},
numpages = {25}
}

@article{jiang2025odverse33,
  author       = {Tianyou Jiang and
                  Yang Zhong},
  title        = {ODVerse33: Is the New {YOLO} Version Always Better? {A} Multi Domain
                  benchmark from {YOLO} v5 to v11},
  journal      = {CoRR},
  volume       = {abs/2502.14314},
  year         = {2025},
  urlREM          = {https://doi.org/10.48550/arXiv.2502.14314},
  doi          = {10.48550/ARXIV.2502.14314},
  eprinttype    = {arXiv},
  eprintREM       = {2502.14314},
  bibsource    = {dblp computer science bibliography, https://dblp.org}
}

@misc{jegham2024yolo,
      title="{YOLO Evolution: A Comprehensive Benchmark and Architectural Review of YOLOv12, YOLO11, and Their Previous Versions}", 
      author={Nidhal Jegham and Chan Young Koh and Marwan Abdelatti and Abdeltawab Hendawi},
      year={2025},
      eprint={2411.00201},
      archivePrefix={arXiv},
      primaryClass={cs.CV},
      url={https://arxiv.org/abs/2411.00201}, 
}

@INPROCEEDINGS{6298404,
  author={Kakar, Pravin and Sudha, N.},
  booktitle={2012 IEEE International Conference on Multimedia and Expo}, 
  title={Authenticating Image Metadata Elements Using Geolocation Information and Sun Direction Estimation}, 
  year={2012},
  volume={},
  number={},
  pages={236-241},
  doi={10.1109/ICME.2012.82},
  ISSN={1945-788X},
  month={July},}

@INPROCEEDINGS{4587784,
  author={Hays, James and Efros, Alexei A.},
  booktitle={2008 IEEE Conference on Computer Vision and Pattern Recognition}, 
  title="{IM2GPS: estimating geographic information from a single image}", 
  year={2008},
  volume={},
  number={},
  pages={1-8},
  doi={10.1109/CVPR.2008.4587784}}

@article{MOHAMMED2025100322,
title = {Architecture review: Two-stage and one-stage object detection},
journal = {Franklin Open},
volume = {12},
pages = {100322},
year = {2025},
issn = {2773-1863},
doi = {https://doi.org/10.1016/j.fraope.2025.100322},
url = {https://www.sciencedirect.com/science/article/pii/S2773186325001100},
author = {Samiyaa Yaseen Mohammed}
}

@INPROCEEDINGS{Black_2022_WACV,
  author={Black, Samuel and Stylianou, Abby and Pless, Robert and Souvenir, Richard},
  booktitle={2022 IEEE/CVF Winter Conference on Applications of Computer Vision (WACV)}, 
  title={Visualizing Paired Image Similarity in Transformer Networks}, 
  year={2022},
  volume={},
  number={},
  pages={1534-1543},
  doi={10.1109/WACV51458.2022.00160}}

@inproceedings{herrmann2024colourbasedgeolocation,
author={Herrmann, Jessica and Bamigbade, Opeyemi and Sheppard, John and Scanlon, Mark},
title="{Perceptual Colour-based Geolocation of Human Trafficking Images for Digital Forensic Investigation}",
booktitle={2024 Cyber Research Conference - Ireland (Cyber-RCI)},
year=2024,
pages={1-8},
month=11,
publisher={IEEE},
doi={10.1109/Cyber-RCI60769.2024.10941203}
}

@InProceedings{joshi2024data,
  title = 	 {Data-Efficient Contrastive Language-Image Pretraining: Prioritizing Data Quality over Quantity},
  author =       {Joshi, Siddharth and Jain, Arnav and Payani, Ali and Mirzasoleiman, Baharan},
  booktitle = 	 {Proceedings of The 27th International Conference on Artificial Intelligence and Statistics},
  pages = 	 {1000--1008},
  year = 	 {2024},
  editor = 	 {Dasgupta, Sanjoy and Mandt, Stephan and Li, Yingzhen},
  volume = 	 {238},
  series = 	 {Proceedings of Machine Learning Research},
  month = 	 {02--04 May},
  publisher =    {PMLR},
  url = 	 {https://proceedings.mlr.press/v238/joshi24a.html}
}

@ARTICLE{bamigbade2025improving,
  author={Bamigbade, Opeyemi and Scanlon, Mark and Sheppard, John},
  journal={IEEE Access}, 
  title={Improving Image Embeddings With Colour Features in Indoor Scene Geolocation}, 
  year={2025},
  volume={13},
  number={},
  pages={79860-79870},
  doi={10.1109/ACCESS.2025.3564496}}

\end{document}